\documentclass[10pt,journal,compsoc, review]{IEEEtran}
\usepackage{amsmath,amsfonts}
\usepackage{algorithmic}
\usepackage{algorithm}
\usepackage{array}
\usepackage[caption=false,font=normalsize,labelfont=sf,textfont=sf]{subfig}
\usepackage{textcomp}
\usepackage{stfloats}
\usepackage{url}
\usepackage{verbatim}
\usepackage{graphicx}
\usepackage{cite}
\usepackage{booktabs}
\usepackage{color}
 
\def\BibTeX{{\rm B\kern-.05em{\sc i\kern-.025em b}\kern-.08em
    T\kern-.1667em\lower.7ex\hbox{E}\kern-.125emX}}
\usepackage{balance}

\begin{document}

\title{FF-ProCams: Feed-Forward Gaussian Splatting for Projector-Camera System}

\author{Ziyao~Wang,
Yuqi~Li,
Wenxing~Zheng, Jiaying~Chen,
and~Chong~Wang
}


\maketitle

\begin{abstract}
Projector-camera (ProCams) systems achieve active scene perception and controllable appearance manipulation via structured illumination, serving as a core infrastructure for spatial augmented reality, projection mapping, and surface reflectance acquisition. Existing inverse-rendering methods for ProCams deliver high-fidelity results but rely on time-consuming per-scene optimization, while mainstream feed-forward 3D reconstruction models produce baked appearance that cannot adapt to spatially varying projector illumination.

To resolve this accuracy–efficiency trade-off, we propose \emph{FF-ProCams}, a \textbf{F}eed-\textbf{F}orward 3D Gaussian inverse-rendering framework for \textbf{ProCams}. A hybrid Mamba2–Transformer encoder aggregates cross-view geometric and photometric cues from sparse multi-view observations, and lightweight heads predict a relightable Gaussian representation in a single forward pass. We further design a projector-aware differentiable renderer to synthesize camera observations under arbitrary active illumination and ProCams poses. To enable feed-forward training, we construct a large-scale synthetic ProCams dataset covering diverse object geometries and surface materials. Experiments show FF-ProCams achieves high-fidelity projector-aware rendering, generalizes to unseen patterns, and supports novel projector–camera poses. Using only 8 input views, it outperforms optimization-based baselines with 297 views, while reducing test-time reconstruction to 0.13 seconds (a three-to-five-order-of-magnitude speedup). The code and data are available at \url{https://github.com/CPREgroup/FF-ProCams}.
\end{abstract}


\section{Introduction}
\IEEEPARstart{P}{rojector--camera}
Projector–camera systems (ProCams) serve as an active 'stimuli' and 'response' framework, which acquire scene information through the collaborative work of projectors and cameras. Different from passive systems that only capture ambient scene information, ProCams actively inject controllable structured illumination into target scenes via projectors~\cite{1_bimber2008visual, 3_grundhofer2018recent, 16_iwai2024projection}, and cameras capture the scene appearance changes induced by encoded projection light. This active illumination mechanism endows ProCams with powerful scene perception and appearance control capabilities and has been widely applied in spatial augmented reality~\cite{15_amano2024enhancing,16_iwai2024projection,17_takeuchi2024projection,18_wang2024vicomp,19_erel2024casper},  projection mapping~\cite{21_ojer2020projection}, and scene reflectance reconstruction fields~\cite{25_chandraker2011duality}.

With the continuous development of projection and imaging hardware, as well as the evolution of neural rendering and 3D scene representation technologies, ProCams-related research has gradually shifted from traditional geometric calibration and photometric correction to high-fidelity 3D reconstruction and controllable simulation. Despite the advances in rendering quality, current ProCams systems still suffer from a prominent practical limitation: they rely heavily on time-consuming scene-specific optimization, which leads to intolerable inference latency and severely hinders deployment in dynamic, interactive, and real-time application scenarios. Meanwhile, mainstream efficient feed-forward 3D reconstruction techniques tailored for passive vision cannot support the active illumination modulation and relighting required by ProCams, resulting in a severe quality-efficiency trade-off in existing methods. Moreover, there is a lack of large-scale datasets that capture appearance responses under varying projection poses and view poses with varying projection patterns, limiting the development of feed-forward reconstruction model for ProCams.

To efficiently estimate the geometry and reflectance of projection scenes and simulate the light reflection and imaging process of ProCams, this paper proposes FF-ProCams, a novel feed-forward 3D Gaussian reconstruction framework oriented to ProCams simulation. Different from previous optimization-based ProCams methods and general passive feed-forward reconstruction models, FF-ProCams takes sparse multiple-view projector images (stimuli) and camera images (response) data as input, and directly predicts complete 3D Gaussian primitives~\cite{42_kerbl20233d} with explicit scene geometry and physical reflection attributes through a single forward inference. This avoids the prohibitive computational cost of iterative optimization, which typically takes minutes or hours per scene, and accelerates the pipeline to merely \emph{0.13 seconds}. Such a three- or four-order-of-magnitude speed improvement makes the multi-view simulation of ProCams feasible for real-time and interactive applications. Our method also produces high-fidelity relighting results by explicitly predicting scene geometry and physical reflection attributes, rather than fixed baked appearance, which supports reliable appearance adaptation to unseen projection patterns and novel projector--camera pose configurations. To facilitate training of this feed-forward paradigm, we construct a synthetic dataset covering diverse geometries, materials, and projection illuminations, which provides sufficient supervised information for geometric reconstruction and reflection estimation. Building on these designs, our framework effectively unites efficient feed-forward inference with high-quality ProCams simulations, achieving promising results in both synthetic and real-world ProCams scenes.

In summary, our main contributions are as follows:

\begin{itemize}
\item We propose FF-ProCams, the \emph{first} dedicated \textbf{F}eed-\textbf{F}orward 3D Gaussian reconstruction framework predicting full \textbf{ProCams} scene representation, including both geometry and reflectance. It eliminates the need for test-time per-scene optimization and realizes realtime prediction of relightable 3D scene representation from sparse multi-view observations.

\item We design a hybrid Mamba2–Transformer Gaussian prediction network, which efficiently aggregates cross-view geometric and photometric cues from observations to achieve accurate prediction of scene geometry and physical appearance-related attributes.

\item We construct a large-scale multi-view ProCams synthetic dataset, providing effective data supervision for feed-forward ProCams reconstruction and projector-aware rendering.
\end{itemize}






\begin{figure*}[t]
    \centering
    \includegraphics[
        width=\textwidth
    ]{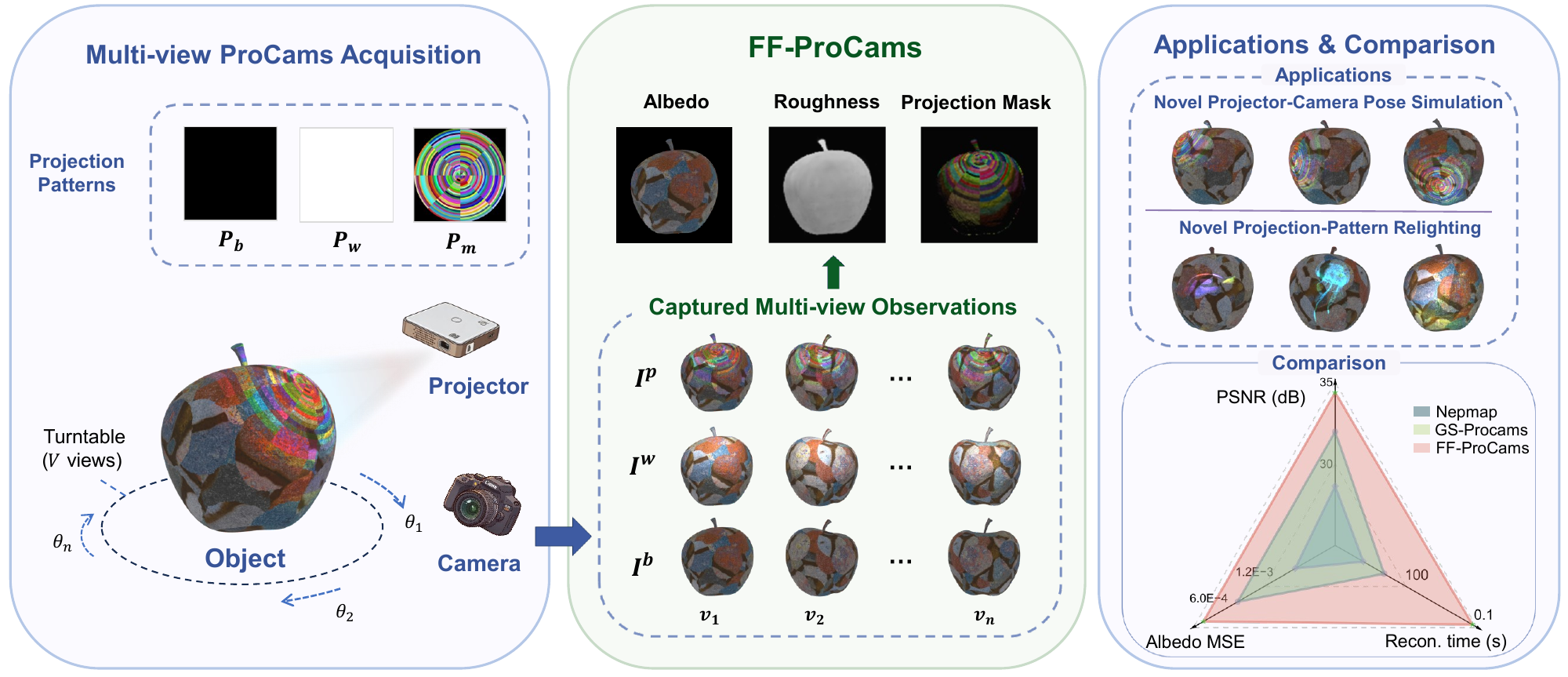}

    \caption{Overview of our FF-ProCams framework. Given sparse multi-view observations captured under black-field, white-field, and active projection patterns, FF-ProCams directly predicts a relightable Gaussian representation with geometric and appearance-related attributes. The resulting representation supports novel projector--camera pose simulation and novel projection-pattern relighting without test-time per-scene optimization.}
    \label{fig:Teaser}
\end{figure*}

\section{Related Work}

This section reviews existing ProCams modeling and 3D neural reconstruction techniques. We organize the techniques into three categories: fixed-pose ProCams simulation, novel-view ProCams simulation, and feed-forward 3D reconstruction methods.

\subsection{Fixed-Pose ProCams Simulation}
Traditional ProCams modeling focuses on characterizes the image-formation relationship among projected illumination, scene geometry, surface appearance, and camera observations under fixed projector--camera poses, aiming to achieve accurate geometric registration ~\cite{sajadi2011autocalibrating, tehrani2019automated, 4_huang2020fast, 56_willi2017robust} and photometric modeling for color reproducing~\cite{sajadi2010adict} and compensation~\cite{60_nayar2003projection, 52_wetzstein2007radiometric, 7_grundhofer2015robust, 2018Practical, 59_kurth2020real, li2023physics}.
Early approaches commonly represented projector-camera light transport using light transport matrices (LTMs), which model the mapping from projector pixels to camera pixels~\cite{49_o2010optical, Yuqi2013}.
Although effective, LTM-based methods generally require dense sampling of projection patterns and the construction, storage, and inversion of large transport matrices. 
Learning-based methods~\cite{9_huang2021deprocams, 23_huang2019end, 10_park2022projector, 40_li2025dpcs,8_huang2021end,li2023extended, 61_kageyama2022online} leverage deep neural networks to implicitly learn the complex radiometric and geometric mapping between projected patterns and camera observations. 
However, they can only achieve view-dependent fitting under determined viewpoints and fail to generalize to variable camera or projector poses.

\subsection{Novel-View ProCams Simulations}
Novel-view adaptive ProCams simulation targets more complex application scenarios where camera or projector poses change, aiming to realize consistent simulation under novel ProCams' poses with unseen projection images. Recent methods incorporate active projector illumination modeling with neural reflectance representations for projector-aware rendering. Specifically, Nepmap~\cite{24_erel2023neural} treats projectors as high-resolution light sources that work with the neural reflectance field~\cite{67_boss2021neural,68_srinivasan2021nerv,69_zhang2021nerfactor,71_boss2021nerd} and jointly optimizes scene and projector parameters from multi-view images with different projection patterns. 
GS-ProCams~\cite{2_deng2025gs} further introduces 2D Gaussian Splatting~\cite{80_Huang2DGS2024} into ProCams simulation, combining explicit Gaussian primitives with projector geometric mapping and BRDF shading to synthesize camera observations under projector illumination. 
The two methods effectively improve projector-aware rendering fidelity, while they still rely on test-time per-scene optimization. Consequently, this per-scene optimization paradigm incurs heavy computational overhead and long convergence times (usually tens of minutes or longer), which limits their practicality for time-critical projection mapping applications. Furthermore, because their underlying scene representations are strictly bound to the specific geometry and material properties captured during training, these methods lack generalizability to unseen environments, requiring a costly retraining process from scratch for every new scene.

\subsection{Feed-Forward Gaussian Reconstruction}

Recently, the emergence of Large Reconstruction Models (LRMs)~\cite{33_hong2024lrm,li2024instant3d,wang2024pf,wei2024meshlrm,xie2024lrm,xu2024dmv3d,38_zhang2024gs,39_ziwen2025long,77_jiang2025anysplat} has offered a promising feed-forward alternative to per-scene optimization for 3D reconstruction. Different from iterative optimization methods, feed-forward models can directly infer renderable 3D scene representations from sparse multi-view observations in a single forward pass, completely avoiding test-time scene optimization and greatly improving reconstruction efficiency. 
Reconstruction models built on Transformers typically tokenize multi-view images into patch sequences and use self attention~\cite{75_vaswani2017attention} to aggregate cross-view information~\cite{33_hong2024lrm,38_zhang2024gs,39_ziwen2025long}.
These feed-forward Gaussian reconstruction methods adopt 3DGS as the output representation and predict Gaussian primitive parameters directly from multi-view images. Specifically, GS-LRM~\cite{38_zhang2024gs} encodes multi-view images into token sequences and decodes pixel-aligned Gaussian parameters, enabling efficient and generalizable 3D reconstruction. Long-LRM~\cite{39_ziwen2025long} further combines Transformer blocks~\cite{75_vaswani2017attention} with Mamba2-based long-sequence modules~\cite{76_dao2024transformers} to process long multi-view token sequences efficiently. 
Together, these methods highlight the importance of modeling long-range
context when aggregating information from multiple views and predicting
Gaussian representations in a feed-forward manner.

However, existing feed-forward Gaussian reconstruction models are primarily designed to reconstruct scene geometry and the appearance observed in the input images for novel-view synthesis. Most existing methods only predict baked scene appearance under fixed ambient illumination, and do not explicitly model the spatially varying active illumination introduced by projectors.
As a result, illumination effects are usually absorbed into the predicted color or SH appearance parameters, yielding Gaussian representations with baked appearance~\cite{36_wang2024crm,37_liu2023one,38_zhang2024gs,39_ziwen2025long,77_jiang2025anysplat}. 
However, this is restrictive for ProCams, where novel projection patterns change the spatially varying incident illumination and require projector-aware 'stimuli-response' modeling. Therefore, such representations
cannot be directly adapted to ProCams. To address the combined challenges of low inference efficiency and limited generalization ability in current methods, this paper proposes a feed-forward ProCams large model, named \emph{FF-ProCams}, that uses Gaussian representations to recover scene geometry and reflectance from sparse multi-view ProCams observations. The following section elaborates the preliminary theoretical basis, formal problem formulation, and the details of our method.

\section{Method}

\begin{figure*}[t]
    \centering
    \includegraphics[
        width=\textwidth
    ]{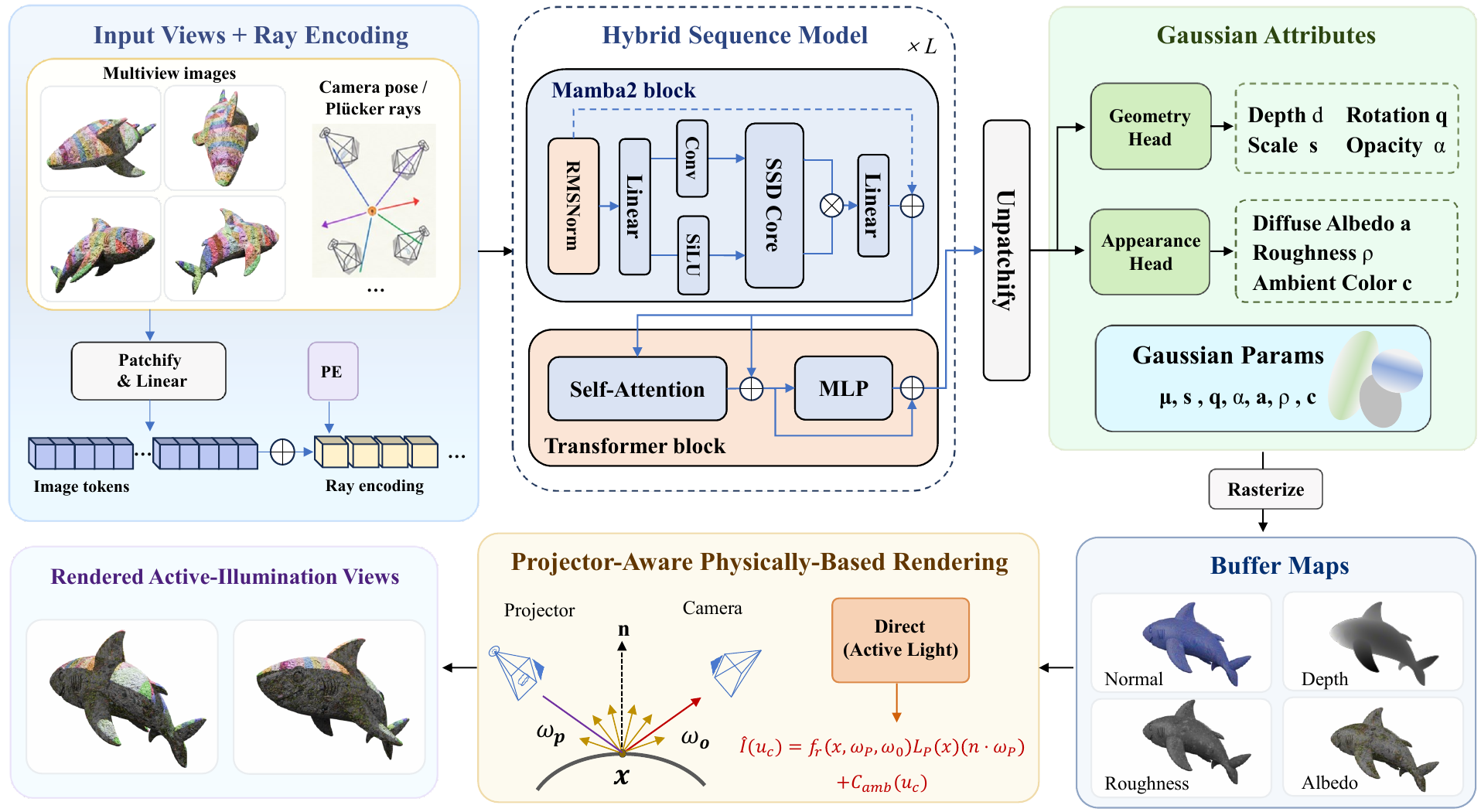}

    \caption{Pipeline of our FF-ProCams framework. The framework encodes multi-view ProCams observations and ray encodings with a hybrid Mamba2--Transformer sequence model, predicts geometry and material Gaussian attributes, and an auxiliary ambient color attribute, rasterizes them into G-buffer maps, and performs projector-aware physically based rendering to synthesize active-illumination views.}
    \label{fig:pipeline}
\end{figure*}

\subsection{Preliminary}

We first introduce the geometric-photometric imaging process of ProCams and the fundamental 3DGS representation used in FF-ProCams, providing unified notation for the subsequent Gaussian attribute prediction and projector-aware rendering modules.

\subsubsection{ProCams Modeling}
\label{sec:projector_camera_modeling}

By integrating active projection with passive camera perception, ProCams capture detailed geometric and photometric properties, enabling robust 3D scene perception and material analysis.

Geometrically, for an arbitrary surface point $\mathbf{x} \in \mathbb{R}^3$ in a 3D scene, its corresponding camera pixel and projector pixel is denoted as $\mathbf{u}_c$ and $\mathbf{u}_p$. The mapping from the 3D surface to the projector plane and camera plane can be defined by the perspective transformations determined by their  intrinsic matrices $\mathbf{K}_{p,c}$ and extrinsic poses $[\mathbf{R}_{p,c} | \mathbf{t}_{p,c}]$ respectively:
\begin{equation}\label{eq:procam_projection}
z_{p,c} \tilde{\mathbf{u}}_{p,c} = \mathbf{K}_{p,c} \left(\mathbf{R}_{p,c} \mathbf{x} + \mathbf{t}_{p,c}\right),
\end{equation}

where $z_{p,c}$ denotes the projection depth and imaging depth, and $\tilde{\mathbf{u}}_{p,c}$ are the homogeneous pixel coordinates of $\mathbf{u}_{p,c}$ respectively.

Photometrically, the surface point $\mathbf{x}$ receives active illumination from the projector, while ambient illumination also contributes to the observed appearance.
Considering these factors, the classical rendering equation describes the outgoing radiance $L_o$ at $\mathbf{x}$ along the viewing direction $\omega_o$ as
\begin{equation}
L_o(\mathbf{x}, \omega_o) =
\int_{\Omega}
f_r(\mathbf{x}, \omega_i, \omega_o)
L_i(\mathbf{x}, \omega_i)
(\mathbf{n} \cdot \omega_i)
d\omega_i ,
\end{equation}
where $\Omega$ denotes the hemisphere around the surface normal, $\omega_i$ is an arbitrary incident light direction, $\mathbf{n}$ is the surface normal, $L_i$ denotes the incident radiance, and $f_r$ is the bidirectional reflectance distribution function (BRDF).

\subsubsection{3D Gaussian Splatting}
Our framework adopts 3D Gaussian Splatting (3DGS)~\cite{42_kerbl20233d} as the explicit geometric base representation of the scene. In standard 3DGS, a 3D scene is parameterized as a point cloud composed of $K$ Gaussian primitives. The spatial distribution of each Gaussian primitive is determined by its center position $\boldsymbol{\mu} \in \mathbb{R}^3$ and a 3D covariance matrix $\boldsymbol{\Sigma}$. To keep the covariance matrix positive semi-definite during gradient-based optimization, $\boldsymbol{\Sigma}$ is explicitly factorized into a diagonal scaling matrix $\mathbf{S}$, constructed from a scaling vector $\mathbf{s} \in \mathbb{R}^3$, and a rotation matrix $\mathbf{R}$ parameterized by a unit quaternion $\mathbf{q} \in \mathbb{R}^4$:
\begin{equation}
\boldsymbol{\Sigma} = \mathbf{R}\mathbf{S}\mathbf{S}^{T}\mathbf{R}^{T},
\end{equation}

In terms of visual appearance, each primitive is also assigned an opacity scalar $\alpha \in \mathbb{R}$ and a color attribute $\mathbf{c}$. The opacity controls its contribution during depth-sorted $\alpha$-blending, 
while $\mathbf{c}$ encodes view-dependent radiance, commonly represented 
with spherical harmonics (SH) coefficients.
Thus, a Gaussian primitive can be represented as
\begin{equation}
\mathcal{G} = \{ \boldsymbol{\mu}, \mathbf{s}, \mathbf{q}, \alpha, \mathbf{c} \},
\end{equation}
The final image is generated through point-based differentiable rasterization. Specifically, 3D Gaussian primitives are first projected onto the 2D image plane. The primitives covering each pixel are then depth-sorted and composited through sequential $\alpha$-blending to compute the final pixel color, enabling efficient and high-fidelity novel-view rendering.

\subsection{Problem formulation}
\label{subsec:problem_formulation}
We consider a standard ProCams setup where the projector and camera are rigidly fixed with known intrinsic and extrinsic pre-calibration, and the object under test is rotated on a turntable to enable full-view coverage. For each rotation angle 
$\theta \in [0,2\pi)$, the system projects patterns and captures the corresponding camera observation. In this setup, the projector serves as a controllable active light source, while the camera captures scene appearance responses under both ambient light and projected patterned light. 

Given the projection patterns and sparse multi-view observations of the ProCams scene, our goal is to
infer a relightable 3D Gaussian representation in a feed-forward manner.
The predicted representation is expected to encode scene geometry and
appearance-related surface attributes, and to support projector-aware rendering under
unseen projection patterns and unseen camera/projector pose configurations
without test-time per-scene optimization. All imaging and projection parameters are fixed during training and inference, ensuring stable geometric and photometric prior constraints for network learning.

The model takes projection patterns and $V$-view ProCams observations with known camera intrinsics and poses as input. For the $v$-th view, the
observations include a black-field image $\mathbf{I}^{b}_{v}\in \mathbb{R}^{H \times W \times 3}$, a white-field
image $\mathbf{I}^{w}_{v}\in \mathbb{R}^{H \times W \times 3}$, and active-projection images
$\mathbf{I}^{p}_{v}\in \mathbb{R}^{H \times W \times 3}$. Here, $H\times W$ denotes the resolution of the captured images. The V-view observations are denoted as
\begin{equation}
\mathcal{I}^{b}=\{\mathbf{I}^{b}_{v}\}_{v=1}^{V}, \quad
\mathcal{I}^{w}=\{\mathbf{I}^{w}_{v}\}_{v=1}^{V}, \quad
\mathcal{I}^{p}=\{\mathbf{I}^{p}_{v}\}_{v=1}^{V}.
\end{equation}

Here $\mathbf{I}^{b}_{v}$ is captured while the projector projects a full-black pattern, providing an observation under minimal projector illumination; $\mathbf{I}^{w}_{v}$ is captured under full-white uniform projection, characterizing the overall surface response to uniform projector illumination;
The active-projection image $\mathbf{I}^{p}_{v}$ is captured under a spatially varying projection pattern 
$\mathbf{P}_{v}$ and provides constraints for projector-aware surface response modeling.
The corresponding camera poses are denoted as
$\mathcal{T}=\{\mathbf{T}_{v}\}_{v=1}^{V}$, where $\mathbf{T}_{v}$
is the pose of the $v$-th camera view.

Based on these inputs, FF-ProCams directly predicts
the Gaussian representation of the scenes:
\begin{equation}
\mathcal{G}=\{\mathcal{G}_{k}\}_{k=1}^{K},
\end{equation}
where $\mathcal{G}$ denotes the predicted set of 3D Gaussian primitives.
Each primitive contains geometric attributes and appearance-related surface
attributes, forming a renderable representation of the ProCams scene. The predicted Gaussian primitives are then rendered to synthesize the target camera observations under a given target camera pose, projector pose, and projection pattern. The detailed rendering formulation is introduced in
Sec.~\ref{subsec:differentiable_render}.

We adopt a fully supervised paradigm with the captured camera images as ground truth. Through the end-to-end differentiable rendering pipeline, loss gradients are back-propagated to update all network parameters and progressively refine the predicted Gaussian primitives.
At inference, the well-trained network directly outputs the complete 3D Gaussian representation for unseen scenes in a single feed-forward pass, eliminating iterative test-time refinement. The inferred Gaussians can be immediately rendered to synthesize views under arbitrary Procams poses and projection patterns, supporting efficient downstream ProCams tasks. 



\subsection{FF-ProCams Overall Architecture}
\label{subsec:FF-ProCams_overall_architecture}


The overall framework of FF-ProCams is illustrated in Fig.~\ref{fig:pipeline}. Given multi-view ProCams observations \(\mathcal{I}^b, \mathcal{I}^w, \mathcal{I}^p\) with known camera intrinsics and poses, FF-ProCams first constructs geometry-aware patch tokens for multi-view sequence encoding. For each input view, image features are augmented with camera-ray encodings derived from the camera intrinsics and poses, forming a geometry-aware feature map that encodes both photometric cues and viewing geometry. 
This feature map is then partitioned into non-overlapping patches, and all pixel features inside each patch are flattened and projected by a linear layer into a \(D\)-dimensional token. The patch tokens from all \(V\) views are concatenated into a multi-view token sequence.

We feed this multi-view token sequence into  a hybrid sequence encoder composed of Mamba2~\cite{75_vaswani2017attention} and Transformer~\cite{76_dao2024transformers} blocks. 
Mamba2 facilitates context propagation over the resulting token sequences, while Transformer blocks explicitly model global correlations among tokens from different views.
With this hybrid encoding strategy, the model aggregates geometric and
photometric cues from multi-view ProCams observations and produces enriched
token representations for subsequent Gaussian attribute prediction.

We augment the original Gaussian representation with additional appearance-related surface attributes for projector-aware rendering.
Specifically, each Gaussian primitive contains geometric attributes, including center $\boldsymbol{\mu}_k$, scale $\mathbf{s}_k$, rotation $\mathbf{q}_k$, and opacity $\alpha_k$, as well as appearance-related attributes, including diffuse albedo $\mathbf{a}_k$, surface roughness $\rho_k$, and an auxiliary ambient color $\mathbf{c}_k$ that models the contribution of ambient illumination to the observed appearance.
Therefore, the $k$-th Gaussian primitive is represented as
\begin{equation}
\mathcal{G}_k =
\{\boldsymbol{\mu}_k, \mathbf{s}_k, \mathbf{q}_k, \alpha_k,
\mathbf{a}_k, \rho_k, \mathbf{c}_k\},
\end{equation}

The predicted physically decoupled Gaussian primitives serve as the core scene representation for projector-aware rendering. The details of the differentiable rendering pipeline will be presented in Section \ref{subsec:differentiable_render}.

\subsection{Gaussian Attribute Prediction Network}

The Gaussian attribute prediction network is the decoding module following the hybrid sequence encoder described in the above section. It takes the context-enhanced token representations as input and maps each token to the attributes of its corresponding Gaussian primitive. Specifically, we employ two lightweight linear prediction heads to generate Gaussian attributes.
The geometry prediction head predicts depth-related geometric attributes including Gaussian depth $d_k$, scale $\mathbf{s}_k$, rotation $\mathbf{q}_k$, and opacity $\alpha_k$, while the appearance prediction head predicts appearance-related surface attributes, including diffuse albedo $\mathbf{a}_k$, surface roughness $\rho_k$, and an auxiliary ambient color $\mathbf{c}_k$. 

We use separate heads to decouple geometric and appearance-related predictions.
The geometric attributes are closely tied to ray-aligned Gaussian placement, visibility, and alpha compositing, whereas the appearance-related attributes are used by the projector-aware renderer to model surface response under active illumination. This separation keeps the prediction module lightweight while allowing different types of Gaussian attributes to be estimated from the shared multi-view token representation.

Since the hybrid sequence encoder has already aggregated geometric and photometric cues across different views and projection patterns, linear token-wise prediction provides a direct mapping from each output token to its corresponding Gaussian primitive, while preserving the ray-aligned correspondence required by feed-forward Gaussian reconstruction.

\textbf{Geometry Prediction Head.} 
Since the spatial position of each Gaussian is directly tied to its corresponding camera ray, we use linear layers to regress the Gaussian depth $d_k$, scale $\mathbf{s}_k$, rotation $\mathbf{q}_k$, and opacity $\alpha_k$ in a token-wise manner. To ensure geometric consistency between the Gaussian center $\boldsymbol{\mu}_k$ and the input-view ray, the Gaussian center \(\boldsymbol{\mu}_k\) is obtained via ray back-projection: 
\begin{equation} 
\boldsymbol{\mu}_k = \mathbf{r}_{o,k} + d_k \mathbf{r}_{d,k}, \end{equation} 
where \(\mathbf{r}_{o,k}\) and \(\mathbf{r}_{d,k}\) denote the origin and direction of the ray corresponding to the \(k\)-th Gaussian primitive, respectively, and \(d_k\) is the network-predicted ray depth. This design constrains each Gaussian center along its corresponding camera ray, thereby preserving the spatial correspondence between the predicted geometry and the input view. 

\textbf{Appearance Prediction Head.} 
For material attribute prediction, we also adopt a token-wise linear head. 
Let \(\mathbf{h}_k\) denote the context-enhanced token associated with the \(k\)-th Gaussian primitive. 
Since cross-view geometric and photometric cues have already been aggregated by the hybrid sequence encoder, the prediction head only needs to read out appearance-related attributes from each token rather than perform additional global reasoning. This lightweight design preserves the one-to-one association between output tokens and Gaussian primitives, and maps each token to the appearance attributes used by the projector-aware renderer:
\begin{equation} 
(\mathbf{a}_k,\rho_k,\mathbf{c}_k) = \phi_m(\mathbf{h}_k), 
\end{equation} 
where \(\phi_m(\cdot)\) denotes the appearance prediction head.

\subsection{Differentiable Projector-Aware Rendering}
\label{subsec:differentiable_render}
Given the Gaussian primitives \(\mathcal{G}\) with the geometric and appearance-related attributes, including the auxiliary ambient color, we design a differentiable projector-aware renderer to synthesize observed images from the predicted attributes by simulating the physical imaging process.

\subsubsection{Attribute Rasterization}
To enable projector-aware rendering under active illumination, we adopt a deferred attribute rendering strategy inspired by DeferredGS~\cite{48_wu2024deferredgs}. 
Given a target camera pose, the predicted Gaussian primitives are first projected onto the target image plane, and differentiable alpha blending is used to generate pixel-aligned G-buffers, include a depth map $\mathbf{Z}$, a normal map $\mathbf{N}$, a diffuse albedo map $\mathbf{A}$, a roughness map $\mathbf{R}$, and an auxiliary ambient color map $\mathbf{C}_{\mathrm{amb}}$.
The appearance-related maps, including diffuse albedo and roughness are obtained by rasterizing the attributes carried by the Gaussian primitives, whereas depth and normals are derived from Gaussian geometric projection and visibility accumulation.

For an arbitrary target camera pixel $\mathbf{u}_c$ and attribute type \(\ell\), let \(\mathcal{K}(\mathbf{u}_c)\) denote the set of Gaussian primitives that cover pixel $\mathbf{u}_c$ and are sorted by depth. The buffered attribute value at pixel $\mathbf{u}_c$ is then given by
\begin{equation}
\hat{\mathbf{B}}^{\ell}(\mathbf{u}_c)
=\sum_{i \in \mathcal{K}(\mathbf{u}_c)}
W_i \, \mathbf{b}_i^{\ell},
\qquad
W_i
=\alpha_i
\prod_{j<i}
\left(1-\alpha_j\right),
\end{equation}
Here, $\mathbf{b}_i^{\ell}$ denotes the $\ell$-th attribute of the $i$-th Gaussian primitive, and $\alpha_i$ denotes its effective opacity. 
The resulting pixel-aligned G-buffers provide the image-space attributes required for the subsequent projector geometric mapping and active-illumination shading.

\subsubsection{Projector Geometry Mapping}

After attribute rasterization, we obtain pixel-aligned G-buffers on the target camera image plane. We then establish the camera-to-projector geometric mapping required for projection-pattern sampling.
Following the perspective projection model introduced in Eq.~\eqref{eq:procam_projection}, we compactly denote the projector and camera projection operators by
\begin{equation}
\Phi_{p,c}(\mathbf{x})
=
\pi\!\left(
\mathbf{K}_{p,c}
\left(
\mathbf{R}_{p,c}\mathbf{x}+\mathbf{t}_{p,c}
\right)
\right),
\label{eq:projection_operator}
\end{equation}
where the subscripts \(p\) and \(c\) denote the projector and camera, respectively, and \(\pi(\cdot)\) denotes perspective division from homogeneous coordinates to two-dimensional image-plane coordinates.

Using the above projection operators, for an arbitrary visible pixel \(\mathbf{u}_c\) in the target camera image, we use the rasterized depth value \(\mathbf{Z}(\mathbf{u}_c)\) to recover the corresponding 3D surface point:
\begin{equation}
\mathbf{x}
=
\Phi_c^{-1}
\left(
\mathbf{u}_c,\mathbf{Z}(\mathbf{u}_c)
\right),
\label{eq:camera_backprojection}
\end{equation}
where \(\Phi_c^{-1}\) denotes the target-camera back-projection operator.
The recovered surface point is then projected onto the projector pattern plane:
\begin{equation}
\mathbf{u}_p
=
\Phi_p(\mathbf{x}),
\label{eq:projector_mapping}
\end{equation}
where \(\mathbf{u}_p\) is subsequently used to sample the projection pattern.

For surface points whose projected coordinates fall outside the valid projector-pattern range, we set the corresponding projection-validity mask to zero to exclude invalid sampling regions from subsequent illumination computation. The resulting
projector coordinates provide the sampling locations required for the subsequent illumination computation.


\subsubsection{Physical Shading and Composition}

After obtaining the projection-pattern sampling coordinate $\mathbf{u}_p$, we follow the illumination modeling strategy of GS-ProCams~\cite{2_deng2025gs} and model the surface appearance using a direct projector-induced reflection component and an ambient color component. 

Specifically, given a projection pattern \(\mathbf{P}_m\), we compute the incident projector illumination by sampling the linearized projection pattern at the mapped projector coordinate, together with the projection-validity mask and geometric attenuation:
\begin{equation}
\mathbf{L}_p(\mathbf{x})
=
M_p(\mathbf{u}_p)\,
\eta_p(\mathbf{x})\,
\mathcal{S}(\widetilde{\mathbf{P}}_m,\mathbf{u}_p),
\end{equation}
where $\widetilde{\mathbf{P}}_m$ denotes the projection pattern converted from
sRGB space to linear intensity space, $\mathcal{S}(\cdot,\mathbf{u}_p)$ denotes bilinear sampling on the projection pattern, $M_p(\mathbf{u}_p)$ is the valid projection mask, and $\eta_p(\mathbf{x})$ accounts for the projector intensity scale and geometric falloff.

To model reflection under projector illumination, similar to GS-ProCams~\cite{2_deng2025gs}, we adopt a simplified Cook--Torrance BRDF with a Lambertian diffuse term and a microfacet specular term~\cite{cook1982reflectance,walter2007microfacet}. As in GS-ProCams, we assume non-metallic dielectric surfaces and fix the normal-incidence Fresnel reflectance to \(F_0=0.04\), rather than predicting a spatially varying specular color.

The predicted roughness is used to control the width of the specular lobe.
Using the G-buffers obtained from attribute rasterization, the pixel-wise BRDF is formulated as

\begin{equation}
f_r(\mathbf{x}_s,\!\omega_p,\!\omega_o)
\!\!=\!\!
\frac{\mathbf{A}(\mathbf{x}_s)}{\pi}
\!\!+\!\!
\frac{
D(\mathbf{n},\!\mathbf{h},\!\rho)\,
F(\omega_o,\!\mathbf{h},\!F_0)\,
G(\mathbf{n},\!\omega_p,\!\omega_o,\!\rho)
}{
4(\mathbf{n}\cdot\omega_p)
(\mathbf{n}\cdot\omega_o)
}.
\label{eq:brdf}
\end{equation}

Here, \(\mathbf{A}(\mathbf{x})\) denotes the diffuse albedo at the visible surface point, and
\(\mathbf{n}=\mathbf{N}(\mathbf{u}_c)\) is the rasterized surface normal at pixel \(\mathbf{u}_c\). 
The half vector is defined as \(\mathbf{h} =(\omega_p+\omega_o)/\|\omega_p+\omega_o\|_2\).
The normal distribution term \(D(\mathbf{n},\mathbf{h},\rho)\) describes the distribution of microfacet normals and is controlled by the predicted roughness \(\rho\).
The Fresnel term \(F(\omega_o,\mathbf{h},F_0)\) is evaluated using the fixed normal-incidence dielectric reflectance.
The geometry term \(G(\mathbf{n},\omega_p,\omega_o,\rho)\) models microfacet masking and shadowing and is also controlled by \(\rho\).

Under active projector illumination, the direct projector-induced reflection observed by the camera is computed as
\begin{equation}
\mathbf{C}_{\mathrm{dir}}(\mathbf{u}_c)
= f_r(\mathbf{x},\omega_p,\omega_o)
\mathbf{L}_{p}(\mathbf{x})
(\mathbf{n}\cdot\omega_p).
\end{equation}
The cosine term $(\mathbf{n}\cdot\omega_p)$ accounts for the influence of the incident angle on the effective illumination intensity. Besides active projector illumination, real ProCams scenes also contain appearance components caused by ambient illumination.
To model this component, the auxiliary ambient color attribute \(\mathbf{c}_k\) is rasterized into a pixel-aligned ambient color map \(\mathbf{C}_{\mathrm{amb}}(\mathbf{u}_c)\). This map represents the
contribution of ambient illumination to the observed appearance.

Therefore, the final rendered color at pixel \(\mathbf{u}_c\) is obtained by combining the projector direct reflection and the ambient color component:
\begin{equation}
\hat{\mathbf{I}}(\mathbf{u}_c)
=
\mathbf{C}_{\mathrm{dir}}(\mathbf{u}_c)
+
\mathbf{C}_{\mathrm{amb}}(\mathbf{u}_c).
\end{equation}

\subsection{Loss Function}
We supervise the predicted 3D Gaussian representation in an end-to-end manner through the projector-aware differentiable rendering process. Given multi-view ProCams observations, the network predicts a set of 3D Gaussian primitives
\(\mathcal{G}=\{\mathcal{G}_k\}_{k=1}^{K}\). The predicted Gaussian representation is then rendered under different projection patterns, providing supervision for geometry reconstruction, appearance-related attribute prediction, and projector-aware relighting.

\textbf{Photometric Loss.}
To supervise image reconstruction under different projection patterns, we combine the \(L1\) loss, structural similarity loss~\cite{wang2004image}, and perceptual loss~\cite{johnson2016perceptual}. 
Following the notation introduced in Sec.~\ref{subsec:problem_formulation}, for the \(v\)-th view, let \(\hat{\mathbf{I}}^{m}_{v}\) denote the rendered image corresponding to the ground-truth observation \(\mathbf{I}^{m}_{v}\), where \(m\in\{b,w,p\}\).
The photometric loss is defined as
\begin{equation}
\mathcal{L}_{\mathrm{img}}^m
= 
(1-\lambda_s)\mathcal{L}_{1}^{m}
+
\lambda_s\mathcal{L}_{\mathrm{SSIM}}^{m}
+
\lambda_{perc}\mathcal{L}_{\mathrm{perc}}^{m},
\label{eq:staged_loss}
\end{equation}
where \(\mathcal{L}_{1}^{m}\) provides pixel-level reconstruction supervision,
\(\mathcal{L}_{\mathrm{SSIM}}^{m}=1-\mathrm{SSIM}(\hat{\mathbf{I}}^{m}_{v},\mathbf{I}^{m}_{v})\) preserves local structural consistency, and \(\mathcal{L}_{\mathrm{perc}}^{m}\) constrains high-level perceptual similarity. All image-level losses are averaged over the training views and image pixels.

To reduce the optimization coupling among geometry, appearance-related surface attributes, and active projector illumination during training, we adopt a staged training strategy: 
We first use black-field supervision to stabilize the geometry and basic surface appearance, then white-field supervision is introduced to constrain
the response to uniform projector illumination, and finally active-projection supervision is added to learn surface responses under spatially varying projector illumination. The staged image supervision term is defined as
\begin{equation}
\mathcal{L}_{\mathrm{img}}
=
\mathcal{L}_{\mathrm{img}}^{b}
+
\mathbb{I}(t>T_w)\mathcal{L}_{\mathrm{img}}^{w}
+
\mathbb{I}(t>T_p)\mathcal{L}_{\mathrm{img}}^{p},
\label{eq:staged_photo_loss}
\end{equation}
where \(t\) denotes the current training iteration, and \(T_w\) and \(T_p\) denote the starting iterations for introducing the white-field supervision and the active projection pattern supervision, respectively.

\textbf{Mask Entropy Loss.}
In our object-centric ProCams setting, the foreground object usually occupies only a portion of the camera image.
When only image reconstruction losses are used, the model may generate redundant Gaussian primitives with non-zero opacity in the background region to fit background colors or local photometric errors, leading to less compact representations and boundary artifacts. To suppress such spurious Gaussians, we introduce a foreground mask loss~\cite{wang2021neus}. Specifically, we adopt a binary cross-entropy form to align the rendered accumulated opacity \(O=\sum_i W_i\) with the foreground mask M:
\begin{equation}
\mathcal{L}_{\mathrm{mask}}
=
-
M\log {O}
-
(1-M)\log(1-{O}).
\label{eq:mask_entropy_loss}
\end{equation}
The foreground mask constraint penalizes opacity outside the target region, encouraging Gaussian contributions to concentrate on the foreground object and reducing boundary artifacts.

\textbf{Geometric Regularization.}
Since Gaussian centers are obtained by back-projecting the predicted depths along camera rays, depth accuracy directly affects the spatial distribution of 3D Gaussians and the accuracy of projector geometric mapping. Moreover, projector-aware shading depends on surface normals and incident light directions, making relighting results sensitive to local geometric noise. Therefore, we regularize the predicted geometry using depth supervision, a depth distribution constraint along camera rays, and surface normal smoothness.

When reference depth is available, we use a depth loss to constrain the predicted depth map:
\begin{equation}
\mathcal{L}_{z}
=
\left\|
\hat{\mathbf{Z}}
-
\mathbf{Z}
\right\|_2^2,
\label{eq:depth_loss}
\end{equation}
where \(\hat{\mathbf{D}}\) and \(\mathbf{D}\) denote the predicted depth map and the reference depth map of a training view, respectively.

Depth supervision alone may still allow multiple Gaussians to be dispersed along the same camera ray, leading to thick or duplicated surfaces. To encourage a compact ray-wise depth distribution, we introduce a depth distortion loss~\cite{80_Huang2DGS2024} that penalizes large depth separations between Gaussians with high rendering contributions:
\begin{equation}
\mathcal{L}_{\mathrm{d}}
=
\sum_{i,j}
W_i W_j
\left|
z_i-z_j
\right|,
\label{eq:dist_loss}
\end{equation}
where \(z_i\) and \(z_j\) denote the depths of the \(i\)-th and \(j\)-th Gaussians along the same pixel ray, and \(W_i\) and \(W_j\) are their alpha-blending weights.

Since projector-aware rendering relies on surface normals to compute incident illumination and BRDF shading, local normal noise can lead to unstable relighting results. We therefore impose a smoothness constraint on the rendered surface normal map to suppress local fluctuations and encourage smoother surface geometry:

\begin{equation}
\mathcal{L}_{n}
=
\left\|
\nabla \hat{\mathbf{N}}
\right\|_1.
\label{eq:normal_smooth_loss}
\end{equation}

\textbf{Roughness Regularization.}
Surface roughness controls the specular-lobe distribution in the microfacet BRDF, but it is not directly supervised by ground-truth material annotations. Without additional constraints, the predicted roughness may absorb local illumination variations or projection-pattern artifacts. To obtain a more stable roughness estimate, we introduce an albedo-guided smoothness constraint~\cite{2_deng2025gs}:
\begin{equation}
\mathcal{L}_{r}
=
\left\|
\nabla \hat{\mathbf{R}}
\right\|_1
\exp
\left(
-
\left\|
\nabla \hat{\mathbf{A}}
\right\|_1
\right),
\label{eq:material_loss}
\end{equation}
where \(\hat{\mathbf{R}}\) and \(\hat{\mathbf{A}}\) denote the rendered
roughness map and diffuse albedo map, respectively. The albedo-gradient
term modulates the roughness smoothness penalty, encouraging smoother
roughness in regions with similar appearance while reducing the penalty
near strong albedo variations.

\begin{figure*}[t]
    \centering
    \includegraphics[
        width=\textwidth
    ]{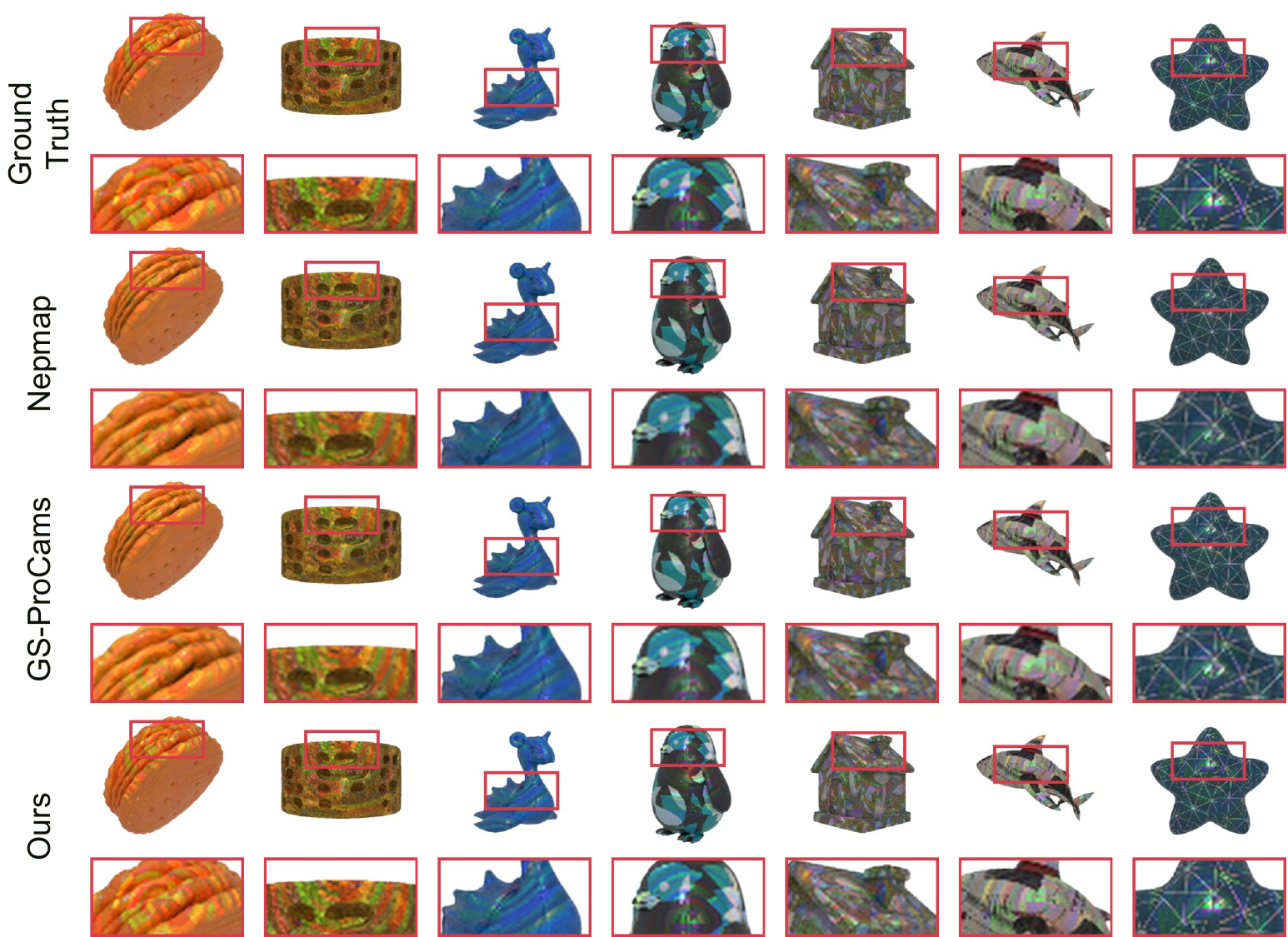}

    \caption{\protect
    Qualitative comparison of ProCams rendering results on the synthetic dataset.  Note that Nepmap and GS-ProCams are optimized using \textbf{297} input views, whereas FF-ProCams uses only \textbf{eight} input views. Compared with the baselines, FF-ProCams more faithfully preserves local color variations, fine projection-pattern details, and the spatial structures of the projected patterns.}
    \label{fig:synthetic_rendering_comparison}
\end{figure*}

\textbf{Overall Loss Function.}
In summary, the training objective of FF-ProCams consists of staged image supervision, mask constraint, geometric regularization, and material regularization. The overall loss function is defined as
\begin{equation}
\begin{aligned}
\mathcal{L}
=
&\mathcal{L}_{\mathrm{img}}
+
\lambda_{\mathrm{m}}\mathcal{L}_{\mathrm{m}}
+
\lambda_{z}\mathcal{L}_{z}
+
\lambda_{\mathrm{d}}\mathcal{L}_{\mathrm{d}}
+
\lambda_{n}\mathcal{L}_{n}
+
\lambda_{r}\mathcal{L}_{r}
\end{aligned}
\label{eq:total_loss}
\end{equation}

By jointly optimizing the staged photometric loss and foreground mask loss with geometric regularization terms and albedo-guided roughness regularization, FF-ProCams learns a feed-forward 3D Gaussian representation that reconstructs multi-view ProCams observations while maintaining projector-aware relighting capability.

\subsection{Training}
During training, the FF-ProCams encoder contains \(L=12\) stacked Mamba2--Transformer hybrid modules, where each module consists of one Mamba2 block and one Transformer block. 

\begin{table}[t] 
    \caption{Quantitative comparison of projector-aware rendering quality and test-time reconstruction time on the synthetic dataset under different numbers of input views. With higher PSNR values, our method is \textbf{three to five orders} of magnitude faster than the baseline method.} \label{tab:synthetic_rendering_comparison}
    \centering \footnotesize \setlength{\tabcolsep}{3.2pt} \renewcommand{\arraystretch}{1.1} 
    \begin{tabular}{lccccc} 
        \toprule Method & Input Views & PSNR $\uparrow$ & SSIM $\uparrow$ & LPIPS $\downarrow$ & Recon. Time $\downarrow$ \\ 
        \midrule Nepmap & 297 & 29.63 & 0.9462 & 0.0530 & $\sim$1 h 30 min \\ 
        Nepmap & 8 & 17.63 & 0.7062 & 0.2530 & $\sim$1 h 30 min \\
        GS-ProCams & 297 & 33.21 & 0.9734 & 0.0230 & 11.16 min \\  
        GS-ProCams & 8 & 16.14 & 0.6675 & 0.2274& 7.15 min \\ 
        Ours & 8 & \textbf{35.76} & \textbf{0.9829} & \textbf{0.0174} & \textbf{0.13 s} \\ \bottomrule 
    \end{tabular} 
\end{table}

 At each training iteration, we use \(V=8\) camera views as multi-view inputs. For each view, we use its black-field image, white-field image, and one active-projection image randomly sampled from the available projection patterns. The active-projection loss is computed with the observation corresponding to the sampled pattern. All images are resized to \(H\times W\), with \(H=W=128\) in our experiments.

For the staged supervision in Eq.~\eqref{eq:staged_loss}, we set
\(T_w=5{,}000\) and \(T_p=10{,}000\) and the model denote the current training iteration by \(t\) and train the model for \(T_{\max}=100{,}000\) iterations.All running times were measured on a workstation equipped with an Intel Core i7-12700K CPU, an NVIDIA RTX 3090 GPU, and 64 GB RAM. AdamW is used as the optimizer with an initial learning rate of \(4\times10^{-5}\).


\section{Experiments}

\begin{figure*}[t]
    \centering
    \includegraphics[
        width=\textwidth
    ]{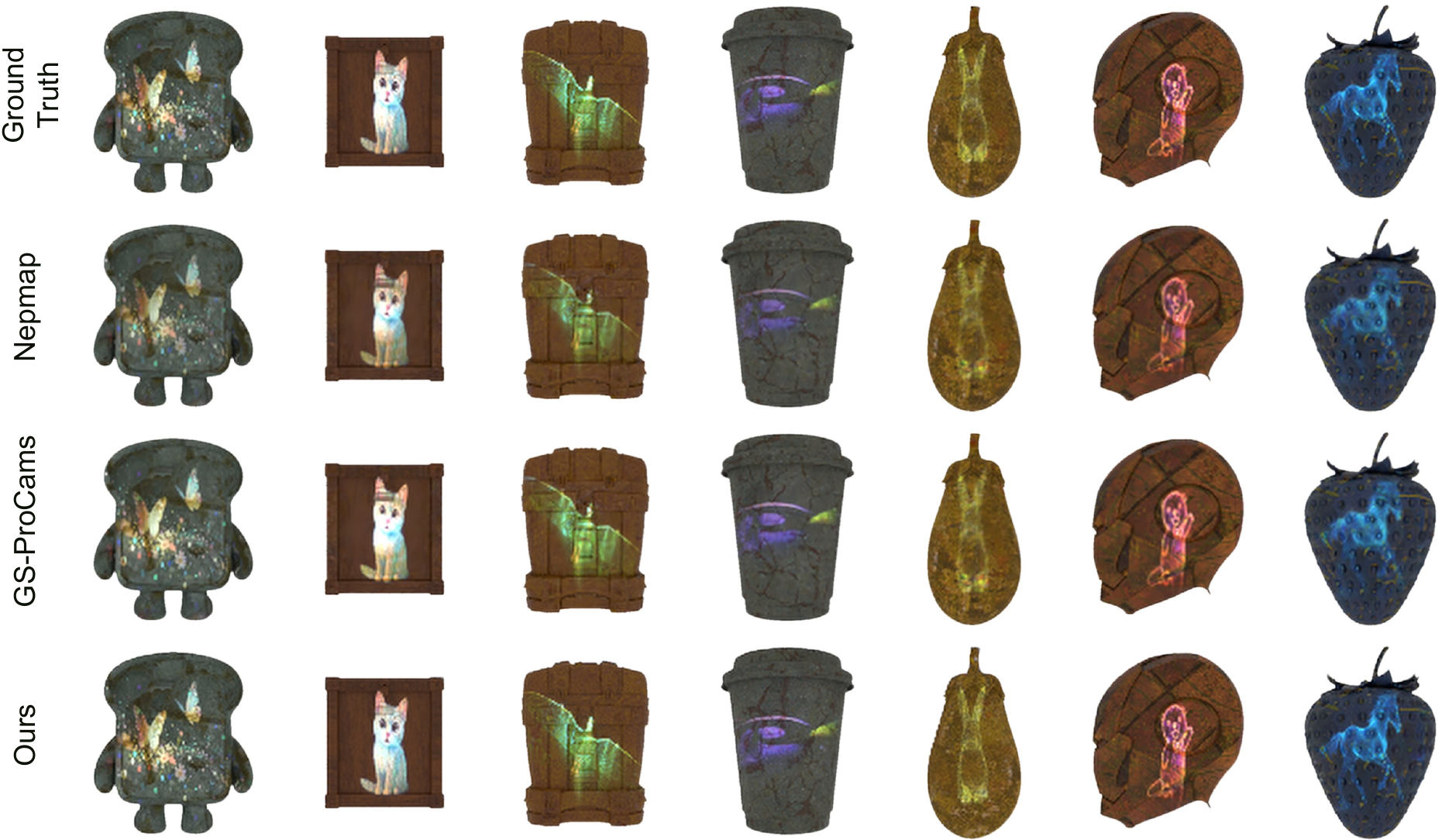}

    \caption{\protect
    Qualitative comparison of novel projection-pattern relighting on the synthetic dataset. FF-ProCams produces projection-pattern structures and color distributions that are more consistent with the ground truth under patterns unseen during training and across diverse object surfaces.}
    \label{fig:novel_pattern_relighting}
\end{figure*}

\begin{figure}[t]
    \centering
    \includegraphics[
        width=\columnwidth
    ]{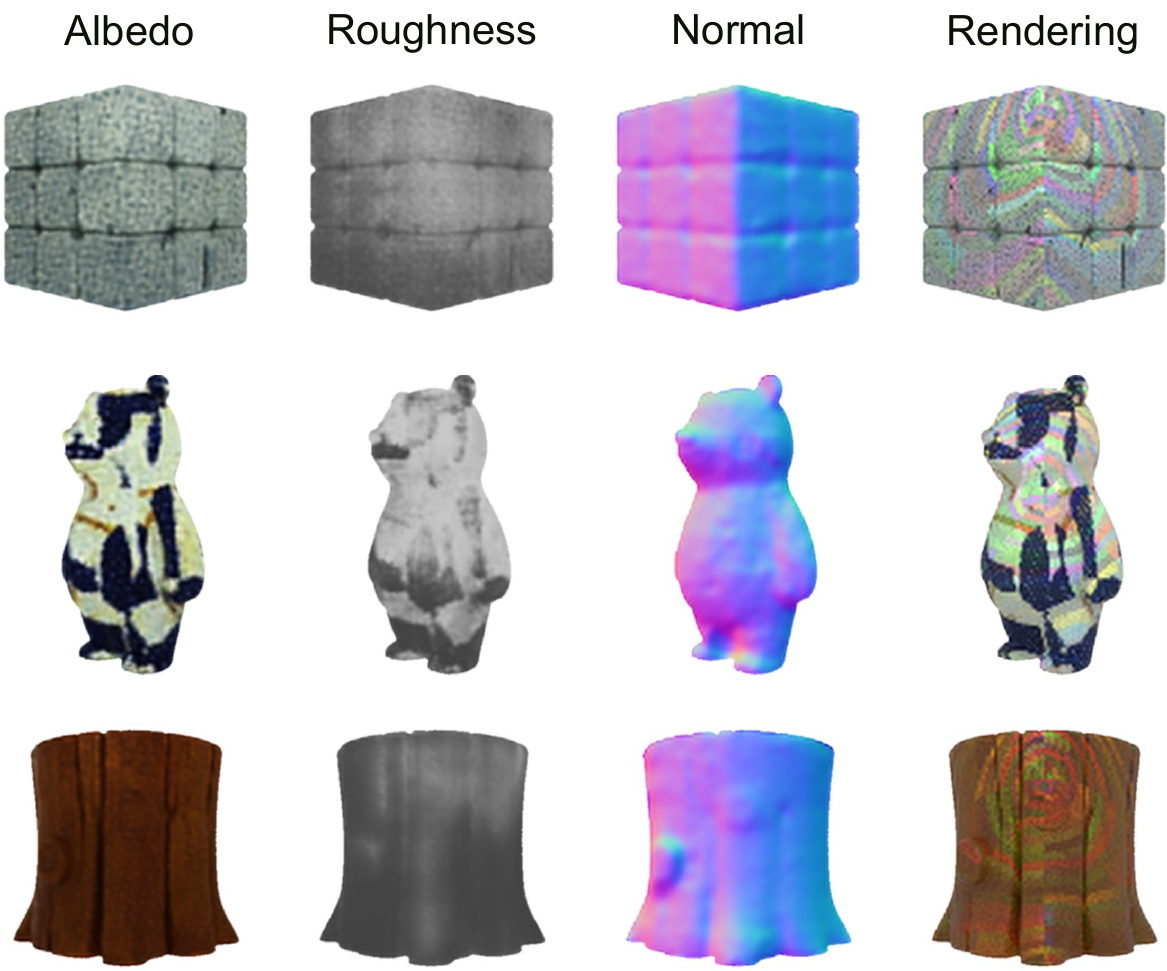}
    \caption{\protect
    Visualization of the predicted appearance-related and geometric attributes on the synthetic test set. Each row corresponds to one test object, with the predicted diffuse albedo, roughness, surface-normal map, and projector-aware rendering shown from left to right. FF-ProCams produces spatially coherent intermediate attributes that support
projector-aware rendering and relighting.}
    \label{fig:synthetic_attribute_visualization}
\end{figure}



\begin{figure*}[t]
    \centering
    \includegraphics[
        width=\textwidth
    ]{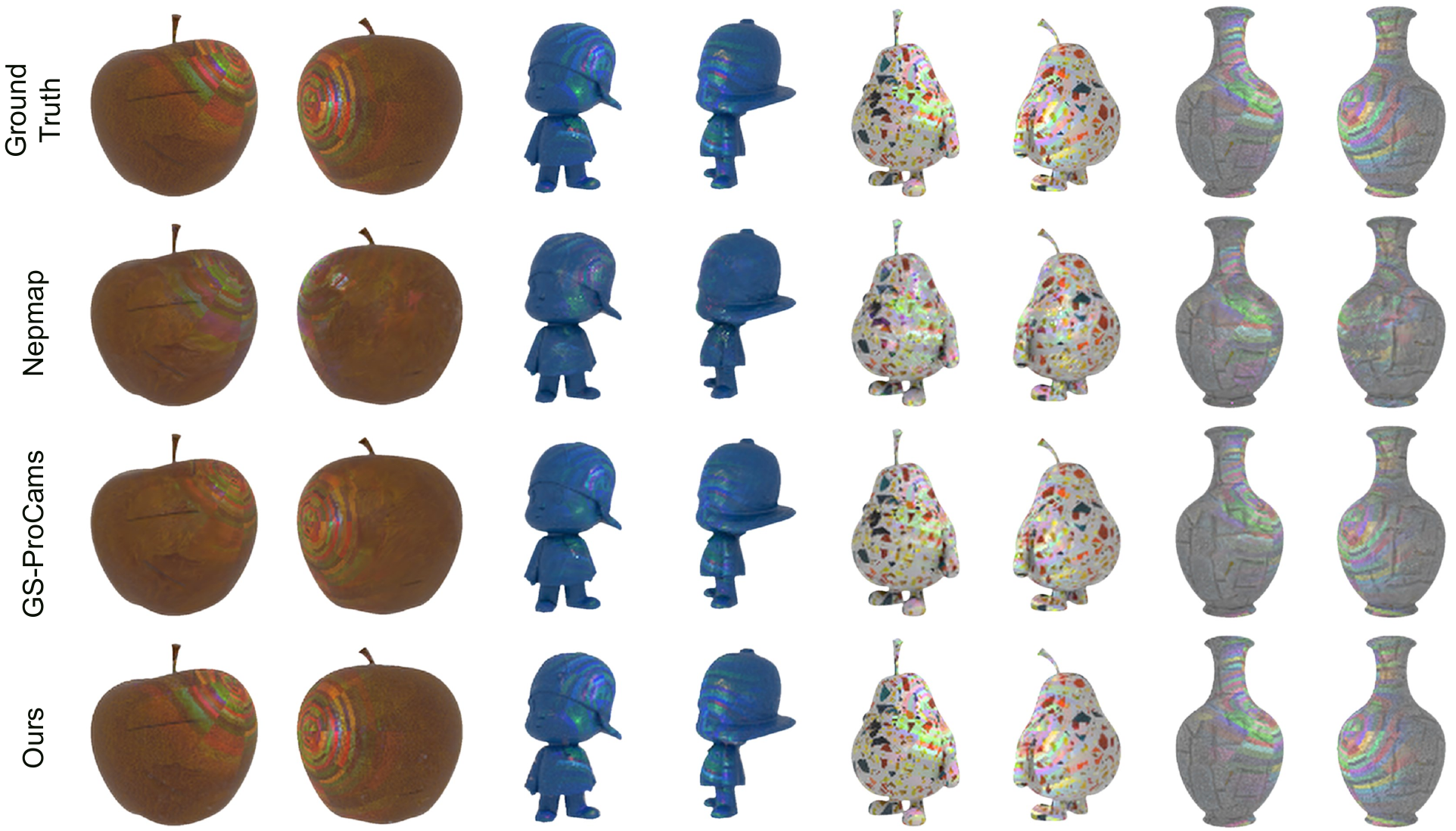}

    \caption{\protect
    Qualitative comparison of ProCams simulation under novel projector--camera poses on the synthetic test set. Note that each pair of columns shows one held-out object under two novel projector--camera pose configurations. Also, both the camera and projector poses and the extrinsic parameters of ProCams are \textbf{different} from those used during training. Compared with the baselines, FF-ProCams produces more geometrically consistent renderings, more accurate projection-pattern alignment, and more stable local appearance.}
    \label{fig:novel_view_relighting}
\end{figure*}

\subsection{Evaluation Settings} 

We evaluated FF-ProCams in terms of rendering quality and test-time efficiency. For rendering quality, we reported Peak Signal-to-Noise Ratio (PSNR), Structural Similarity Index Measure (SSIM), and Learned Perceptual Image Patch Similarity (LPIPS)~\cite{zhang2018unreasonable} between the rendered images and the ground-truth observations under different projection patterns or projector--camera
pose configurations. All image-quality metrics were averaged over all test objects, target views, and corresponding evaluation conditions.

For efficiency evaluation, we reported the test-time reconstruction time of each method, which was defined as the time required to obtain a renderable scene representation and render the target-view image under the specified evaluation condition. 
For the optimization-based baselines, the reported time corresponded to the per-scene optimization and rendering process.
For FF-ProCams, it included feed-forward Gaussian prediction and projector-aware rendering, and was averaged over all test scenes.

\begin{table}[t] 
    \caption{Quantitative comparison of novel projection-pattern relighting on the synthetic dataset. The two compared methods use 297 views, while FF-ProCams use only eight views.} \label{tab:novel_pattern_relighting} \centering \footnotesize \setlength{\tabcolsep}{14pt} 
    \renewcommand{\arraystretch}{1.1} 
    \begin{tabular}{lccc} 
        \toprule Method & PSNR $\uparrow$ & SSIM $\uparrow$ & LPIPS $\downarrow$ \\ 
        \midrule Nepmap & 28.58 & 0.9363 & 0.0638 \\ 
        GS-ProCams & 30.21 & 0.9584 & 0.0398 \\ 
        Ours & \textbf{32.36} & \textbf{0.9611} & \textbf{0.0349} \\ 
        \bottomrule 
    \end{tabular} 
\end{table}

\subsection{Experiments on Synthetic Dataset}

\textbf{Dataset.}
To address the lack of large-scale ProCams data for feed-forward training, we constructed a Blender-based synthetic dataset under controlled projector--camera conditions.
We used Hunyuan3D~\cite{78_zhao2025hunyuan3d} to generate object meshes and applied physically based material maps from MatSynth~\cite{79_vecchio2024matsynth} to the generated meshes. 
This process produced ProCams scenes with varied object geometries and surface appearances.
In the synthetic ProCams setup, the relative pose between the camera and the projector was fixed, and multi-view observations were obtained by rotating the object. 
Following Neural Projection Mapping (Nepmap)~\cite{24_erel2023neural}, we used lollipop patterns as spatially varying active-projection signals. For each object and each view, we rendered one black-field image, one white-field image, and six active-projection images under different lollipop patterns.

In addition to RGB observations, the dataset stored the projection patterns, foreground masks, depth maps, camera intrinsics and poses, and projector intrinsics and extrinsics. Each object contained observations from eight views.

The synthetic dataset contained 43 physically based materials and 150 object meshes. By applying each material to each mesh, we obtained 6,450 geometry--material ProCams scenes. To evaluate generalization to unseen object geometries, we split the dataset at the mesh level, using 105 meshes for training, 15 meshes for validation, and 30 meshes for testing. All material variants of the same mesh were assigned to the same set to avoid geometry leakage across training, validation, and testing.
The synthetic test set was used for ProCams rendering comparison, novel projection-pattern relighting, and novel projector--camera pose simulation experiments.

\begin{table}[t] 
    \caption{Quantitative comparison of novel projector--camera pose simulation on the synthetic dataset. The two compared methods use 297 views, while FF-ProCams use only eight views.} \label{tab:novel_views_relighting} \centering \footnotesize \setlength{\tabcolsep}{14pt} \renewcommand{\arraystretch}{1.1} 
    \begin{tabular}{lccc} 
        \toprule Method & PSNR $\uparrow$ & SSIM $\uparrow$ & LPIPS $\downarrow$ \\ 
        \midrule Nepmap & 25.85 & 0.9079 & 0.0875 \\ 
        GS-ProCams & 27.84 & 0.9263 & 0.0532 \\ 
        Ours & \textbf{32.49} & \textbf{0.9575} & \textbf{0.0308} \\ 
        \bottomrule 
    \end{tabular} 
\end{table}

\textbf{ProCams Rendering Comparison.}
We compared FF-ProCams with Nepmap~\cite{24_erel2023neural} and GS-ProCams~\cite{2_deng2025gs} on the synthetic test set. Both baselines optimized each test scene separately. 
In contrast, FF-ProCams directly predicted a 3D Gaussian representation with geometric and appearance-related attributes from sparse multi-view ProCams observations in a single forward pass. 
The projector-aware differentiable renderer used this representation to synthesize images under active projector illumination without test-time per-scene optimization.
As reported in Table~\ref{tab:synthetic_rendering_comparison}, FF-ProCams consistently outperformed Nepmap and GS-ProCams across PSNR, SSIM, and LPIPS under the same eight-view setting. 
Notably, despite using only eight input views, FF-ProCams also surpassed the dense-view variants of both baselines using 297 views. Moreover, its feed-forward prediction reduced the test-time reconstruction time to \(0.13\) seconds, whereas the optimization-based baselines required several minutes or hours. 
This substantial efficiency improvement resulted from the feed-forward prediction paradigm.

As shown in Fig.~\ref{fig:synthetic_rendering_comparison}, the zoomed-in regions highlight local projection-pattern details under active projector illumination. Nepmap preserved the overall object appearance but tended to smooth out fine projected structures. GS-ProCams retained sharper details than Nepmap, but exhibited local color deviations and less accurate projection-pattern structures on complex surfaces. 
In contrast, FF-ProCams more faithfully preserved local color variations and the spatial structures of the projected patterns. 
These qualitative results were consistent with the quantitative comparison and demonstrated the effectiveness of FF-ProCams in reproducing active-illumination appearance.

To inspect the intermediate attributes used by the projector-aware renderer, we visualized the predicted diffuse albedo, roughness, and surface-normal maps. 
As shown in Fig.~\ref{fig:synthetic_attribute_visualization}, the albedo maps captured the main diffuse surface colors, the roughness maps exhibited spatially coherent reflectance variations, and the surface-normal maps provided plausible surface orientation cues. These buffers provided the surface information required for projector-aware shading and relighting.



\begin{figure}[t]
    \centering
    \includegraphics[
        width=\columnwidth
    ]{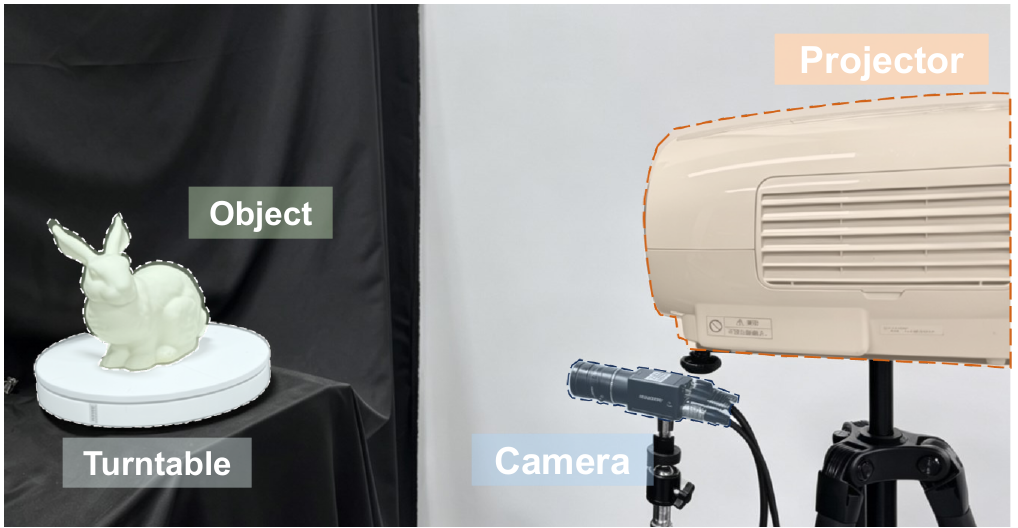}
    \caption{\protect
    Real-world ProCams acquisition setup. The system consists of a geometrically calibrated projector--camera pair and a turntable-mounted object. Multi-view observations are captured by rotating the object while keeping the projector and camera stationary with a fixed relative pose.}
    \label{fig:real_setup}
\end{figure}

\begin{figure}[t]
    \centering
    \includegraphics[
        width=\columnwidth
    ]{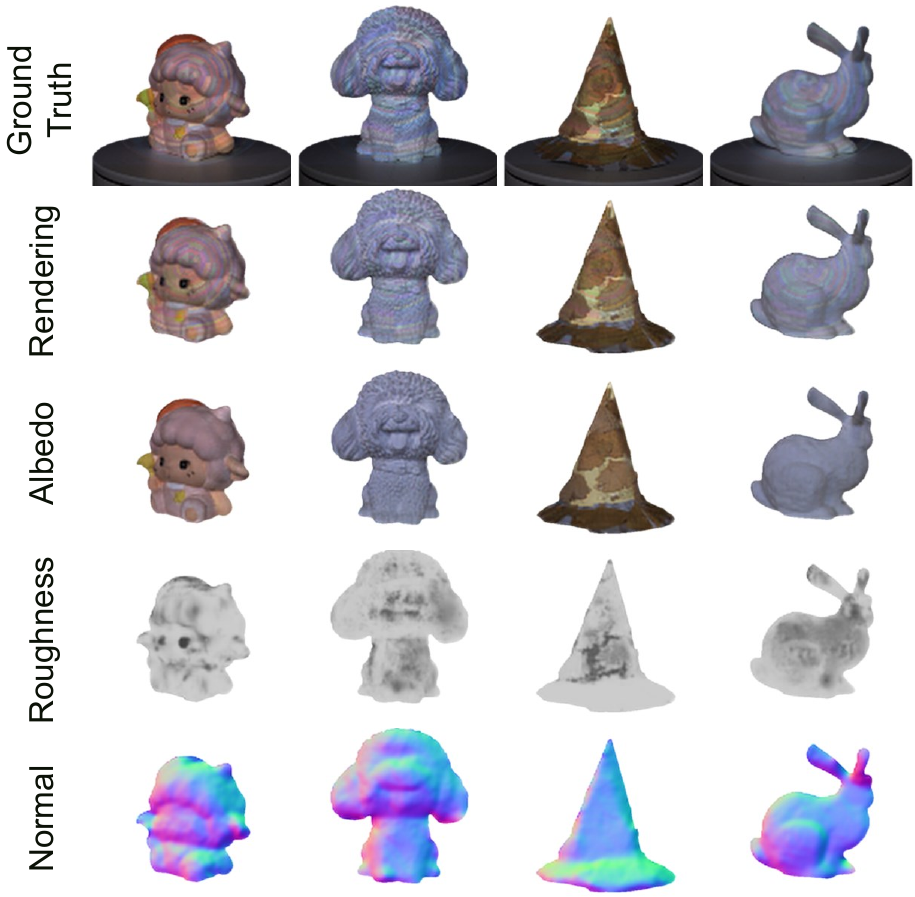}
    \caption{\protect
    Real-world rendering and attribute prediction results of FF-ProCams on held-out test objects. FF-ProCams reconstructs active-illumination appearances that closely match the captured ground-truth observations and predicts spatially coherent roughness, diffuse albedo, and surface-normal maps from sparse real-world ProCams observations.}
    \label{fig:real_reconstruction_results}
\end{figure}

\textbf{Novel Projection-Pattern Relighting.} 
To evaluate novel projection-pattern relighting, we tested FF-ProCams using projection patterns that were excluded from training. This experiment examined whether the predicted Gaussian representation could synthesize surface appearance responses under unseen projection patterns across different object surfaces.

As reported in Table~\ref{tab:novel_pattern_relighting}, FF-ProCams achieved the best performance across all three metrics and outperformed the compared baselines under unseen projection patterns. 
Fig.~\ref{fig:novel_pattern_relighting} further provides qualitative comparisons on different object surfaces under unseen projection patterns. 
FF-ProCams preserved the spatial structure and color distribution of the projection patterns more closely than the baselines.

\textbf{Novel Projector--Camera Pose Simulation.}
We further evaluated projector-aware simulation under novel projector--camera pose configurations, where both the camera and projector poses differed from those in the training observations. This setting was more challenging than changing only the projection pattern because the model needed to maintain geometric consistency, projector geometric mapping, and stable surface appearance under the novel pose configurations.

As reported in Table~\ref{tab:novel_views_relighting}, FF-ProCams outperformed Nepmap and GS-ProCams in rendering quality under this setting.
As shown in Fig.~\ref{fig:novel_view_relighting}, FF-ProCams better preserved object geometry, projection-pattern alignment, and local appearance under the novel camera and projector poses. 
Its rendered results remained close to the ground truth in both overall appearance and active-projection details. 
In contrast, the baselines exhibited greater geometric inconsistency and less stable projection-pattern alignment.
These results indicated that the predicted relightable Gaussian representation supported novel projector--camera pose simulation without test-time per-scene optimization.

\begin{figure*}[t]
    \centering
    \includegraphics[
        width=\textwidth
    ]{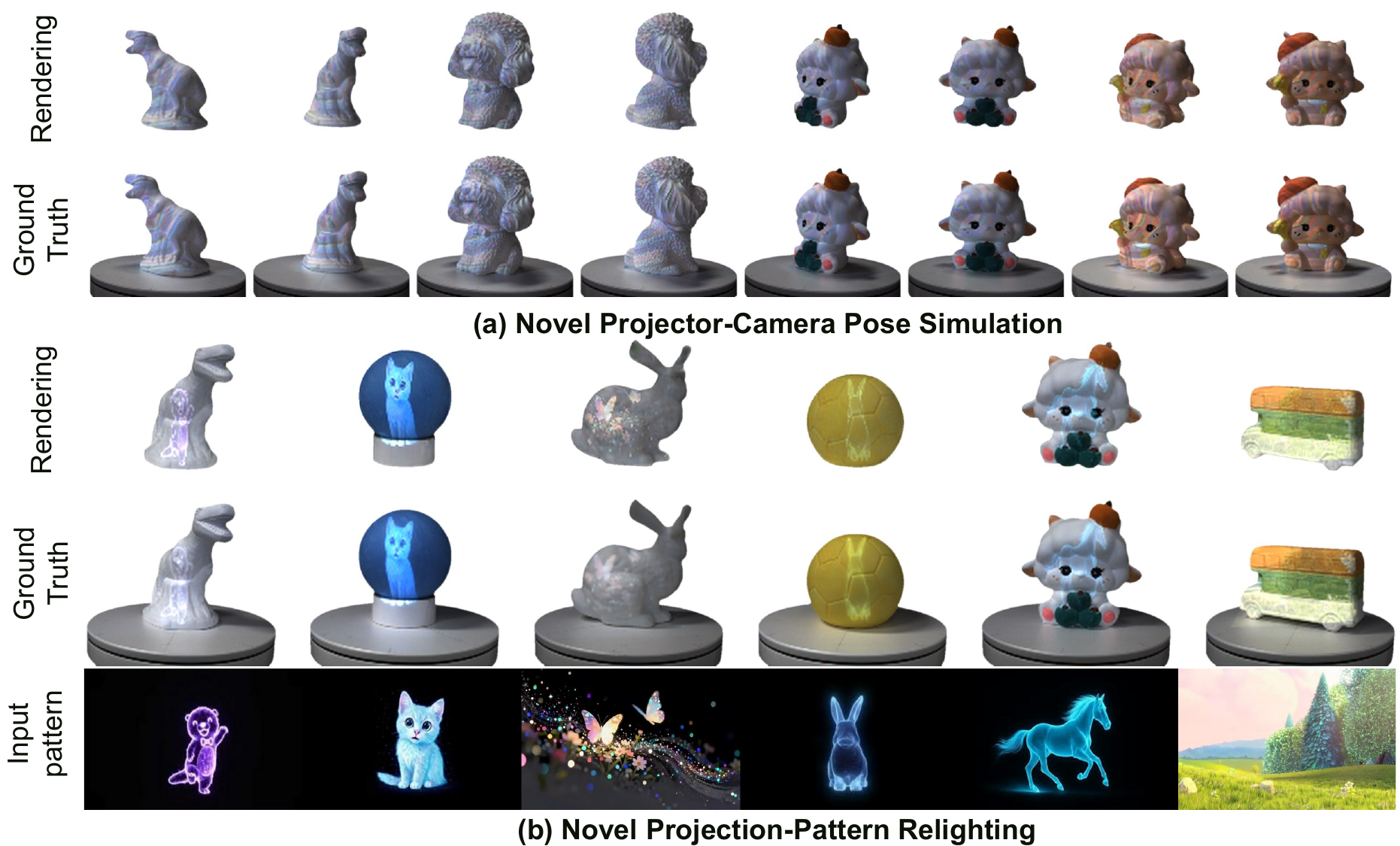}

    \caption{\protect
    Real-world rendered results of FF-ProCams. (a) Novel projector--camera pose simulation. Novel projector--camera pose simulation under separately calibrated configurations whose relative poses \textbf{differed} from those used during real-world fine-tuning. The rendered results are compared with the corresponding captured ground-truth observations. (b) Novel projection-pattern relighting using projection patterns unseen during real-world fine-tuning, with the corresponding input patterns shown in the bottom row. FF-ProCams produces plausible active-illumination appearances with consistent projection-pattern alignment, preserved projection-pattern structure, and stable surface appearance on real-world objects.}
    \label{fig:real_world_relighting_results}
\end{figure*}

\subsection{Experiments on Real-World Dataset}
To validate the applicability of FF-ProCams in real-world ProCams scenarios, we built a standard ProCams acquisition setup consisting of a Hikvision industrial camera, an Epson projector, and a target object mounted on a
turntable, as shown in Fig.~\ref{fig:real_setup}. 
During acquisition, the geometrically calibrated camera--projector pair remained at a fixed relative pose, while the target object was rotated to capture multi-view observations. The camera captured images at a resolution of \(512\times512\), whereas the projector displayed patterns at a resolution of \(1920\times1080\).

We collected data from 40 real-world objects with diverse geometries and surface appearances.
Following the acquisition protocol used for the synthetic dataset, we captured eight views and each view contained one black-field image, one white-field image, and six active-projection images captured under different lollipop patterns.
We performed an object-level split, using 25 objects for training and 15 objects for testing.
To adapt FF-ProCams to real-world imaging conditions with limited real data, we first pretrained the model on the synthetic dataset and then fine-tuned it on the 25 real-world training objects. All real-world evaluations were conducted on the 15 held-out test objects, which were not used during fine-tuning.

For the novel projector--camera pose evaluation, we additionally captured ground-truth observations under calibrated projector--camera configurations whose relative poses differed from those used during real-world fine-tuning.
During testing, we separately evaluated novel projection-pattern relighting and novel projector--camera pose simulation. 
Since ground-truth annotations for the predicted appearance-related attributes were unavailable, the real-world evaluation focused on rendering and relighting quality rather than direct attribute-estimation accuracy.

\textbf{Real-World Rendering Results.}
We first evaluated FF-ProCams on real-world ProCams observations. As shown in Fig.~\ref{fig:real_reconstruction_results}, the model synthesized active-illumination appearances using only eight input views. The rendered results preserved the overall appearance of the object, projection-induced color variations, and object boundaries across real objects with various geometries and surface appearances. 
In addition to the final rendered images, FF-ProCams predicted appearance-related and geometric buffers, including diffuse albedo, roughness, and surface-normal maps.  
The predicted albedo maps exhibited spatially coherent color distributions while largely suppressing projection-induced appearance variations.
The roughness maps showed smooth reflectance-related variations, while the surface-normal maps captured plausible local surface-orientation changes. 
Together with the rendered results, these qualitative observations suggested that FF-ProCams could recover a relightable Gaussian representation from sparse real-world ProCams observations.

We already show in Table~\ref{tab:synthetic_rendering_comparison} that Nepmap and GS-ProCams exhibited substantial performance degradation when their input views were reduced from 297 to 8 on the synthetic benchmark. 
Moreover, their acquisition manner relies on dense observations captured under varying camera and static projector poses, whereas our real-world setup acquires only eight views by rotating the object while keeping the calibrated projector--camera pair at a fixed relative pose. 
Because of this substantial difference in both input density and acquisition manner, we did not conduct a direct real-world comparison with these baselines. Instead, the real-world experiments focused on evaluating the generalization of FF-ProCams to sparse, standard fixed-pair ProCams acquisition and its ability to support rendering and relighting on held-out objects.

\textbf{Real-World Relighting Results.} 
We further validated the simulation capability of FF-ProCams on real-world ProCams data. As shown in Fig.~\ref{fig:real_world_relighting_results}(a), we first evaluated novel projector--camera pose simulation. In this setting, both the camera and projector were repositioned relative to the target object, resulting in a projector--camera relative pose that differed from that of the input observations. 
FF-ProCams produced plausible active-illumination appearances under the changed projector--camera pose configurations, while maintaining stable object appearance and consistent projection-pattern alignment with the captured ground-truth observations.

As shown in Fig.~\ref{fig:real_world_relighting_results}(b), we further evaluated novel projection-pattern relighting. In this setting, FF-ProCams was evaluated using projection patterns that were unseen during real-world fine-tuning. 
The model synthesized plausible relighting results under these unseen patterns and preserved the main spatial structures of the unseen projection patterns
across real objects with diverse geometries and surface appearances.

Compared with the synthetic setting, real-world simulation was affected by sensor noise, calibration inaccuracies, projector response nonlinearity, and more complex surface reflectance. Therefore, quantitative results were not given in the real-world experiments.
Overall, these consistent visual results indicated that FF-ProCams could generalize to real-world ProCams acquisition conditions after fine-tuning and support both novel projector--camera pose simulation and novel projection-pattern relighting on held-out real objects.

\begin{table*}[t]
    \centering
    \caption{Ablation study of different Mamba2--Transformer configurations. M and T denote Mamba2 and Transformer blocks, respectively, and the numbers preceding M and T indicate the numbers of consecutive corresponding blocks. The notation $\times n$ denotes repeating the pattern $n$ times. The all-T and all-M variants use only Transformer and Mamba2 blocks, respectively. Here, the symbol '$-$' indicates that evaluation cannot be performed because training does not converge with this architecture. }
    \label{tab:ablation_memory_quality}
    \footnotesize
    \setlength{\tabcolsep}{14pt}
    \renewcommand{\arraystretch}{1.1}
    \begin{tabular}{lcccccc}
        \toprule
        Method 
        & Train Mem.
        & Eval Mem.
        & Recon. Time
        & PSNR $\uparrow$
        & SSIM $\uparrow$
        & LPIPS $\downarrow$\\
        \midrule
        all-T
        & 19.3G
        & 6.4G
        & 0.43s
        & \textbf{35.83}   
        & 0.9735   
        & \textbf{0.0153} \\
        $\{3\mathrm{M}1\mathrm{T}\}\times6$ 
        & 18.1G    
        & 5.1G   
        & 0.11s   
        & 29.23   
        & 0.9361  
        & 0.0739 \\
        $\{7\mathrm{M}1\mathrm{T}\}\times3$
        & 17.9G     
        & \textemdash 
        & \textemdash 
        & \textemdash 
        & \textemdash 
        & \textemdash \\
        all-M 
        & 17.7G  
        & \textemdash 
        & \textemdash 
        & \textemdash 
        & \textemdash 
        & \textemdash \\
        Ours($\{\mathrm{MTMT}\}\times6$) 
        & 18.4G 
        & 5.3G 
        & 0.13s 
        & 35.76 
        & \textbf{0.9829} 
        & 0.0174 \\
        \bottomrule
    \end{tabular}
\end{table*}

\begin{figure}[t]
    \centering
    \includegraphics[
        width=\columnwidth
    ]{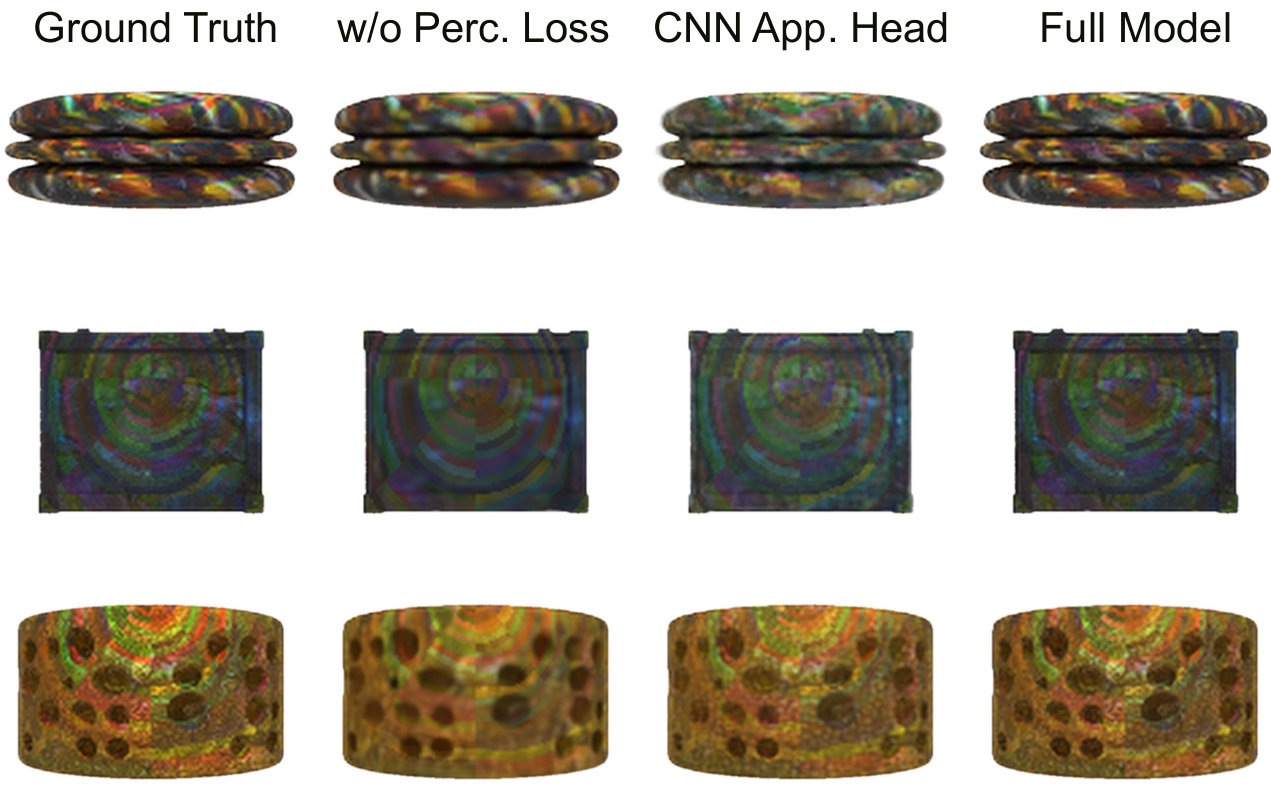}
    \caption{\protect
    Qualitative ablation results of the perceptual loss and appearance head design. Removing the perceptual loss leads to blurrier textures and weaker high-frequency projection-pattern details, while the CNN-based appearance head produces greater local color deviations and oversmoothing. In contrast, the full model preserves sharper appearance structures and more accurate projection-induced color variations.}
    \label{fig:ablation_perceptual_cnn}
\end{figure}

\subsection{Ablation Study}
\textbf{Effect of Network Architecture.} 
We first evaluated different Mamba2--Transformer configurations to analyze the trade-off between rendering quality and computational efficiency. 
As reported in Table~\ref{tab:ablation_memory_quality}, the all-Transformer variant achieved slightly better PSNR and LPIPS than our hybrid design, but it required higher training memory and a substantially longer test-time reconstruction time. 
Increasing the proportion of Mamba2 blocks reduced memory consumption and reconstruction time.
However, the \(\{3\mathrm{M}1\mathrm{T}\}\times6\) variant suffered a substantial degradation in rendering quality despite achieving the shortest reconstruction time.
The \(\{7\mathrm{M}1\mathrm{T}\}\times3\) and all-M variants failed to converge stably, and therefore their evaluation memory, reconstruction time, and rendering-quality metrics were not reported.

Compared with the all-Transformer variant, our \(\{\mathrm{MTMT}\}\times6\) design reduced the test-time reconstruction time from \(0.43\) s to \(0.13\) s while achieving the highest SSIM and comparable overall rendering quality. These results indicated that the proposed hybrid architecture provided a more favorable balance between cross-view modeling capability and computational efficiency.

\begin{table}[t]
    \centering
    \caption{Ablation study of the perceptual loss and appearance head design.}
    \label{tab:ablation_perceptual_cnn}
    \footnotesize
    \setlength{\tabcolsep}{12pt}
    \renewcommand{\arraystretch}{1.1}
    \begin{tabular}{lccc}
        \toprule
        Method 
        & PSNR $\uparrow$ 
        & SSIM $\uparrow$ 
        & LPIPS $\downarrow$ \\
        \midrule
        w/o Perc. Loss 
        & 32.36 
        & 0.9627 
        & 0.0645 \\ 
        CNN App. Head
        & 27.69 
        & 0.9160 
        & 0.0667 \\
        Full Model 
        & \textbf{35.76}
        & \textbf{0.9829} 
        & \textbf{0.0174} \\
        \bottomrule
    \end{tabular}
\end{table}

\textbf{Effect of Perceptual Loss and Appearance Head Design.} 
We further ablated the perceptual loss and the appearance head design, as reported in Table~\ref{tab:ablation_perceptual_cnn}. 
Removing the perceptual loss degraded all three evaluation metrics, especially LPIPS. This result indicated that pixel-wise reconstruction losses alone did not provide sufficient supervision for perceptually faithful appearance reconstruction under active projection. 
As shown in Fig.~\ref{fig:ablation_perceptual_cnn}, the model without perceptual supervision produced blurrier textures and weaker projection-induced high-frequency details. 
In contrast, the full model recovered sharper appearance structures and more faithful color variations, resulting in better PSNR, SSIM, and LPIPS.

We also compared the proposed token-wise linear appearance head with a CNN-based appearance prediction head. 
This comparison was motivated by the common use of CNN-based decoders in dense prediction tasks, where local convolutional operations provide image-plane spatial priors and can facilitate local texture reconstruction. 
We therefore examined whether such local convolutional prediction was beneficial for Gaussian appearance attribute prediction in feed-forward ProCams reconstruction.

As reported in Table~\ref{tab:ablation_perceptual_cnn}, the token-wise linear appearance head consistently outperformed the CNN-based head in PSNR, SSIM, and LPIPS. 
As further shown in Fig.~\ref{fig:ablation_perceptual_cnn}, the linear head preserved sharper appearance details and more faithful projection-pattern responses. 
One possible explanation is that the Mamba2--Transformer encoder had already aggregated multi-view geometric and photometric information, allowing each token to be directly mapped to the appearance-related attributes of its corresponding Gaussian primitive.
In contrast, the CNN-based head introduced additional local mixing on the image-plane token grid. Such mixing may combine tokens associated with different depths, object boundaries, or illumination responses, potentially weakening primitive--observation correspondence and smoothing high-frequency active-illumination details. These results suggested that direct token-wise appearance attribute prediction was more suitable for FF-ProCams.


\section{Limitations and Conclusion}

\subsection{Discussion}
Despite the favorable performance of FF-ProCams in sparse-view ProCams reconstruction and relighting, there are still three major practical limitations that limit its applicability, which are discussed as follows.

\textbf{Resolution.} Constrained by GPU memory capacity and the computational overhead of our feed-forward model, the current framework adopts an input and output resolution of 
$128 \times 128$ for both training and inference. Although this setting achieves competitive rendering quality and ultra-high inference efficiency, it inevitably sacrifices fine-grained details to some extent. For high-resolution projection mapping applications that demand pixel-level detail fidelity, the performance of the current version still has room for further improvement. This bottleneck can be alleviated in future work via compact network redesigns and GPU  upgrades to support higher-resolution training.

\textbf{The Number of Views.} Our method currently takes eight sparse views as standard input, which strikes a balance between reconstruction accuracy and acquisition efficiency, and this configuration already outperforms optimization-based baselines under the same sparse input. Increasing the number of views can improve rendering fidelity but will raise the input computational cost and higher GPU memory consumption. Meanwhile, the model’s reconstruction robustness degrades when facing extremely sparse input views. Therefore, the current approach is only for conventional sparse-view ProCams scenarios and requires further optimization for extreme sparse observation conditions.

\textbf{The Fixed Poses of Procams}
In the current implementation, the relative poses between the ProCams system and the target scene remain fixed throughout model training. Empirically, we found that training with arbitrary, unconstrained ProCams relative poses leads to unstable model convergence and severe performance degradation. Therefore, our method is only applicable to fixed-pose ProCams acquisition setups, and its generalization ability for flexible, arbitrary pose configurations needs further enhancement.

\subsection{Conclusion}

We present FF-ProCams, a feed-forward 3D Gaussian reconstruction framework that enables efficient, end-to-end ProCams reconstruction and relighting without per-scene test-time optimization.
By combining a hybrid Mamba2–Transformer encoder, physically decoupled Gaussian attributes, and projector-aware differentiable renderer, FF-ProCams delivers high-fidelity active-illumination rendering under unseen patterns and novel poses. Critically, its feed-forward paradigm cuts per-scene reconstruction to 0.13 seconds, resolving the long-standing efficiency bottleneck of conventional ProCams methods. In addition, the constructed large-scale ProCams synthetic dataset provides effective data support for feed-forward ProCams research. Extensive synthetic and real-world experiments validate its superior rendering quality, generalization, and efficiency.
Future work will focus on higher-resolution reconstruction, robustness to extremely sparse views, and generalization to arbitrary projector–camera pose configurations.

\bibliographystyle{IEEEtran}
\bibliography{egbib}

\newpage

\vfill

\end{document}